\documentclass[final]{cvpr}

\usepackage{times}
\usepackage{epsfig}
\usepackage{graphicx}
\usepackage{amsmath}
\usepackage{amssymb}

\usepackage[table]{xcolor}
\usepackage{xcolor}
\usepackage[utf8]{inputenc} 
\usepackage[T1]{fontenc}    
\usepackage{url}            
\usepackage{booktabs}       
\usepackage{amsfonts}       
\usepackage{nicefrac}       
\usepackage{microtype}      
\usepackage[ruled,vlined]{algorithm2e}
\definecolor{ceruleanblue}{rgb}{0.2, 0.35, 0.68}
\usepackage[bottom]{footmisc}

\newcommand{\uu}{\mathbf{u}}
\newcommand{\iu}{\mathrm{u}}
\newcommand{\I}{\mathrm{I}}
\newcommand{\p}{\mathrm{p}}
\newcommand{\x}{\mathrm{x}}
\newcommand{\m}{\mathrm{m}}
\newcommand{\mi}{\mathbb{I}}
\newcommand{\ent}{\mathbb{H}}
\newcommand{\loss}{\mathcal{L}}
\newcommand{\TC}{\mathrm{\scriptscriptstyle TC}}
\newcommand{\A}{\mathrm{\scriptscriptstyle A}}

\newcommand{\J}{\mathrm{\scriptscriptstyle J}}

\DeclareMathOperator*{\argmax}{arg\,max}
\DeclareMathOperator*{\argmin}{arg\,min}

\usepackage[pagebackref=true,breaklinks=true,colorlinks,bookmarks=false]{hyperref}

\def\cvprPaperID{2557} 
\def\confYear{CVPR 2021}
\begin{document}

\title{DyStaB: Unsupervised Object Segmentation via Dynamic-Static Bootstrapping\thanks{Also ArXiv:2008.07012, August 16, 2020}
}

\author{Yanchao Yang\thanks{Equal contributions. Our implementation and trained models are available at: https://github.com/blai88/unsupervised\_segmentation}\\
Stanford University\\
{\tt\small yanchaoy@cs.stanford.edu}
\and
Brian Lai\footnotemark[2]\\
UCLA Vision Lab\\
{\tt\small b4lai@g.ucla.edu}
\and
Stefano Soatto\\
UCLA Vision Lab\\
{\tt\small soatto@ucla.edu}
}

\maketitle

\begin{abstract}
We describe an unsupervised method to detect and segment portions of images of live scenes that, at some point in time, are seen moving as a coherent whole, which we refer to as objects.
Our method first partitions the motion field by minimizing the mutual information between segments. Then, it uses the segments to learn object models that can be used for detection in a static image. Static and dynamic models are represented by deep neural networks trained jointly in a bootstrapping strategy, which enables extrapolation to previously unseen objects. While the training process requires motion, the resulting object segmentation network can be used on either static images or videos at inference time.
As the volume of seen videos grows, more and more objects are seen moving, priming their detection, which then serves as a regularizer for new objects, turning our method into unsupervised continual learning to segment objects. 
Our models are compared to the state of the art in both video object segmentation and salient object detection. In the six benchmark datasets tested, our models compare favorably even to those using pixel-level supervision, despite requiring no manual annotation.
\end{abstract}

\vspace{-0.2cm}
\section{Introduction}

\begin{figure}[!t]
  \centering
  \includegraphics[width=0.47\textwidth]{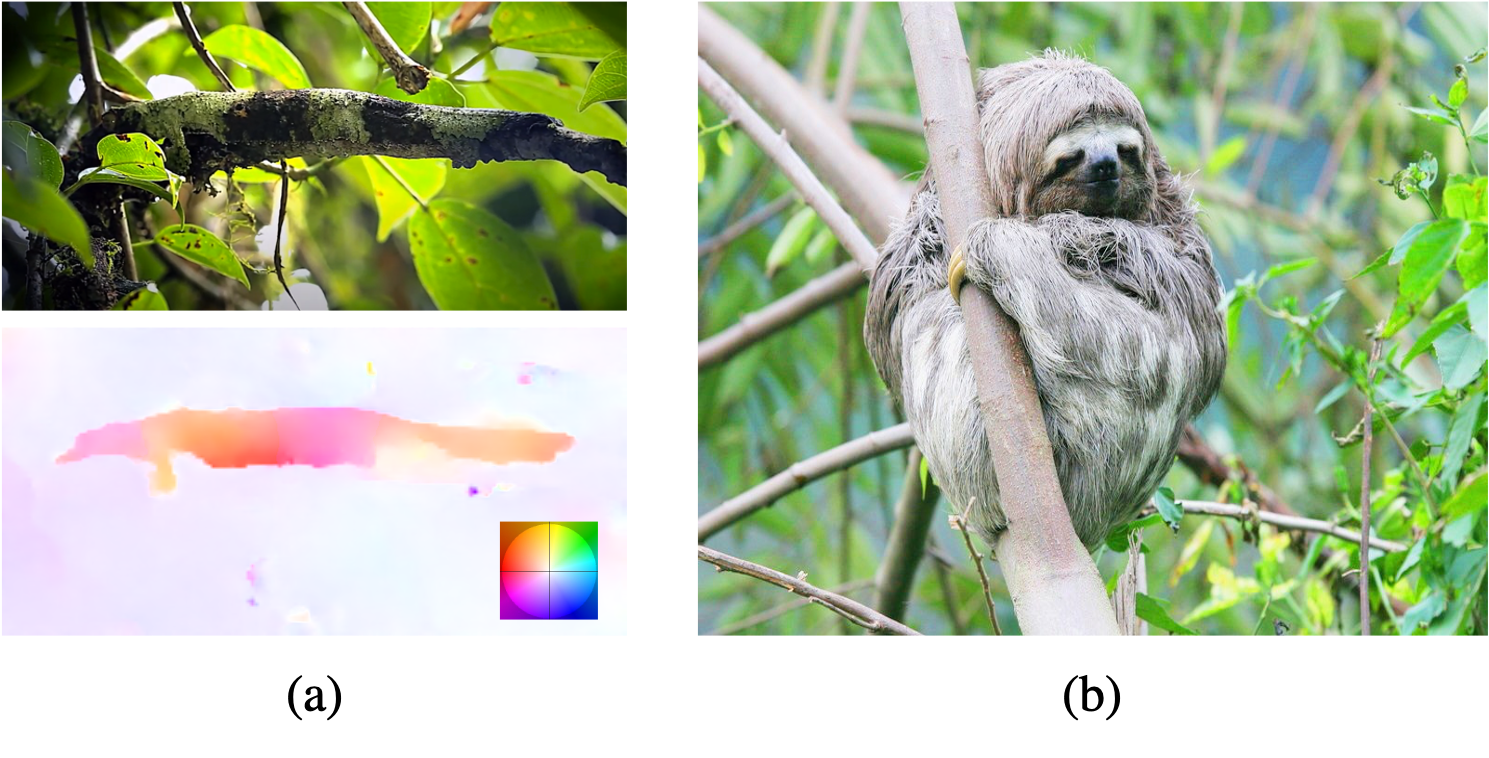}
  \vspace{-0.1cm}
  \caption{Dynamic-static bootstrapping. (a) A lizard is hard to detect when still thanks to camouflage (top left). However, it is easy to see once it moves (bottom left; optical flow visualized using the inset color wheel). Once learned the lizard, a never-before-seen sloth (b) can be easily detected in a static image, exploiting the static model learned from the moving lizard.}
  \vspace{-0.2cm}
  \label{fig:lizard-sloth}
\end{figure}

The ability to segment the visual field into coherently moving regions is among the traits most broadly shared among visual animals \cite{albright1995visual,gibson2014ecological,goldstein2016sensation}. 
Even camouflaged, moving objects are easy to spot (Fig. \ref{fig:lizard-sloth}). 
During early development, humans spend considerable amounts of time interacting with a single moving object before losing interest \cite{bambach2018toddler}, which may help prime object models and learn invariances \cite{spelke1990principles}.
In contrast, the mature visual system can learn an object with a few static examples; objects do not need to move in order to be detected. This suggests using motion as a cue to bootstrap object models that can be used for detection in static images, with no need for explicit supervision. 
Objects that have never been seen moving are considered part of whatever background they are part of, at least until they move. 
As time goes by, more and more objects are seen moving, thus improving one's ability to detect and segment objects in static images (Tab.~\ref{tab:chi-improves}). 
The more objects are bootstrapped in a bottom-up fashion, the easier they are to detect top-down, priming better motion discrimination, which in turn results in more accurate object detection. 
This synergistic loop gradually improves both the diversity of objects that can be detected and the accuracy of the detection.

\begin{figure*}[!t]
  \centering
  \includegraphics[width=0.85\textwidth]{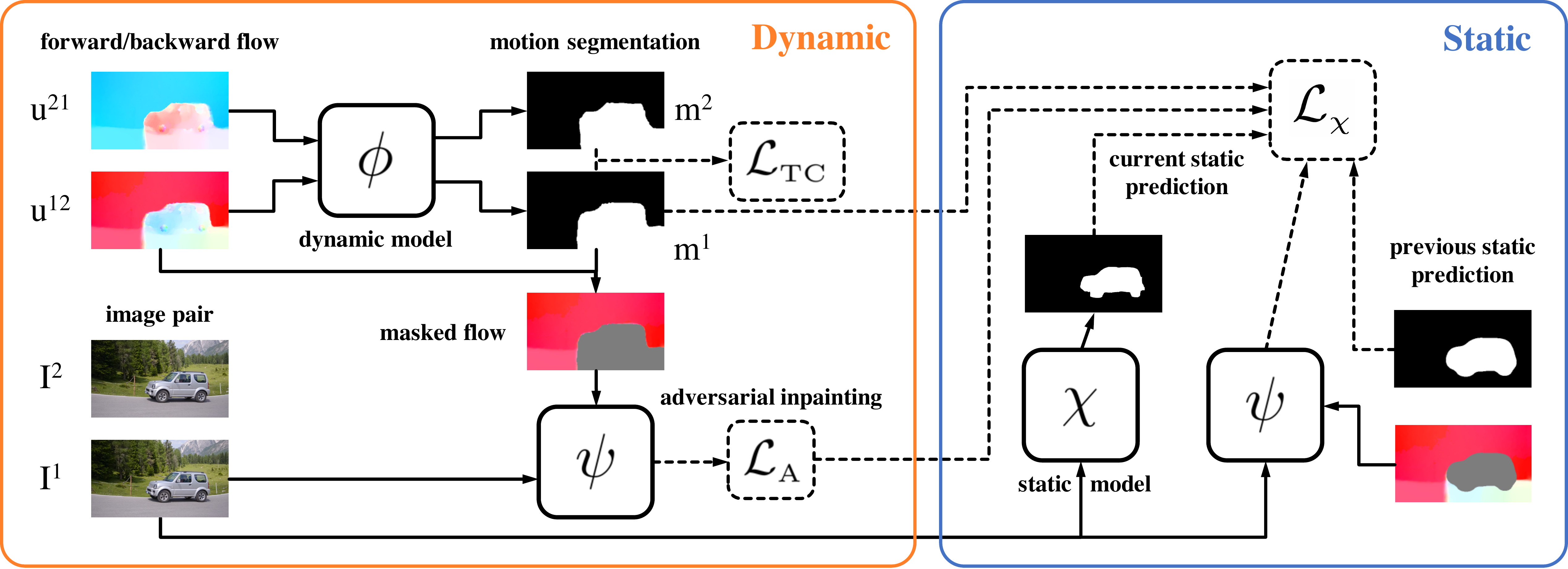}
  \vspace{-0.0cm}
  \caption{System overview. {\color{orange} Dynamic}: motion segmentation model described in Sec.~\ref{subsec:CIS_TC_SP}; {\color{ceruleanblue} Static}: object model described in Sec.~\ref{subsec:static_object}. The training (``dynamic-static bootstrapping'' in Sec.~\ref{subsec:joint_percept}) iterates between these two models. In particular, once $\chi$ (static object model) is trained, it can be used as a ``top-down'' object prior to bias the motion segmentation network $\phi$ (dynamic model) in a feedback loop. $\psi$ is the adversarial inpainting network that enforces minimal mutual information between motion field partitions (two $\psi$'s are identical), and losses are represented using dashed lines/boxes.}
  \vspace{-0.2cm}
  \label{fig:system}
\end{figure*}

We present a method to learn object segmentation using unlabeled videos, that at test time can be used for both moving objects in videos and static objects in single images. 
The method uses a motion segmentation module that performs temporally consistent region separation. The resulting motion segmentation primes a detector that operates on static images, and feeds back to the motion segmentation module, reinforcing it. We call this method Dynamic-Static Bootstrapping, or {\em DyStaB}.
During training, the dynamic model minimizes the mutual information between partitions of the motion field, while enforcing temporal consistency, which yields a state-of-the-art unsupervised motion segmentation method. Putative regions, along with their uncertainty approximated during the computation of mutual information in the dynamic model, are used to train a static model. 
The static model is then fed back as a regularizer in a top-down fashion, completing the synergistic loop.

One might argue that every pixel in the image back-projects to something in space that we could call an object. 
However, the training data biases the process towards objects that exist at a scale that is detectable relative to the size of the pixel and the magnitude of the motion. For instance, individual leaves in an outdoor video might not be seen at a resolution that allows detecting them as independent objects. Instead, the tree may be detected as moving coherently. 
So, the definition of objects is conditioned on the training data and, in particular, the scale and distribution of their size and relative motion.

With this caveat, our contribution is two-fold: First, a deep neural network trained with unlabeled videos that achieves state-of-the-art performance in motion segmentation. It exploits mutual information separation and temporal consistency to identify candidate objects. Second, a deep neural network to perform object segmentation in single images, bootstrapped from the first. The static model uses as input both the output of the dynamic model and its uncertainty, to avoid self-learning. The two models are trained jointly in a synergistic loop. The resulting object segmentation models outperform the state of the art by 10\% in average precision in both video and static object segmentation across six standard benchmarks. Despite not requiring any manual annotation, our method also outperforms recent supervised ones by almost 5\% on average.

\begin{figure*}[!t]
	\centering
	\begin{minipage}[t]{.55\textwidth}
		\centering
		\includegraphics[trim={0.55cm 0.0cm 0 0},clip,width=.95\linewidth]{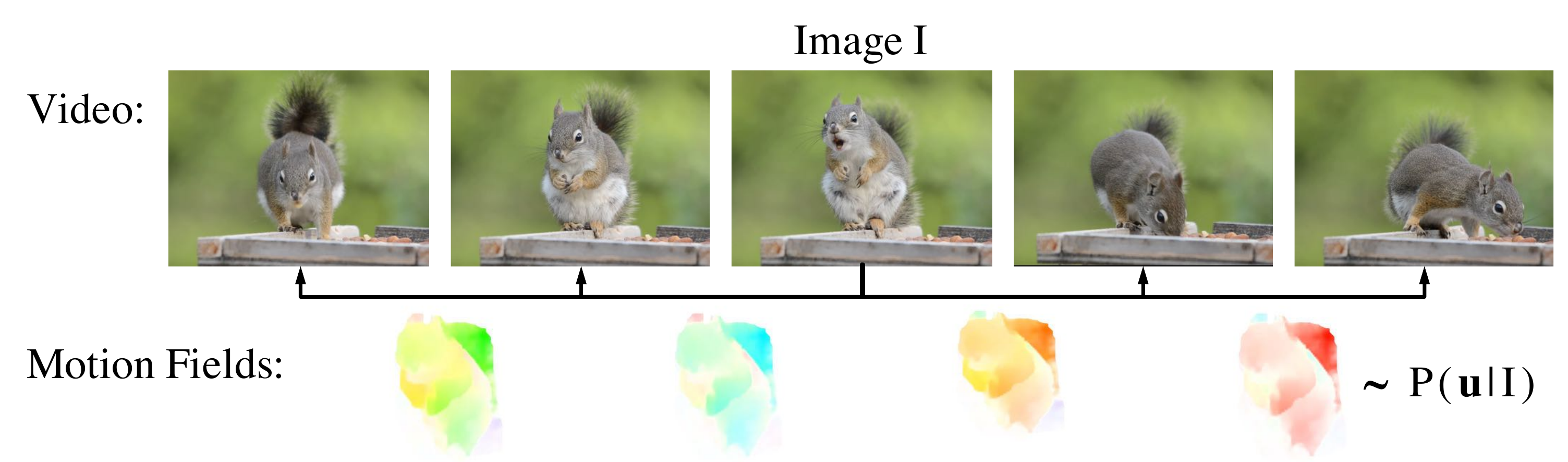}
		\vspace{-0.0cm}
		\caption{A single image $\I$, renders possible motion fields $\uu$ of an independently moving object as if they are sampled from the conditional distribution $\p(\uu|\I)$. Thus, given the image, one can complete partial observations of the flow fields in Fig.~\ref{fig:over-under}.}
		\vspace{-0.2cm}
		\label{fig:conditional}
	\end{minipage}
	\hspace{0.2cm}
	\begin{minipage}[t]{.33\textwidth}
		\centering
		\includegraphics[width=.75\linewidth]{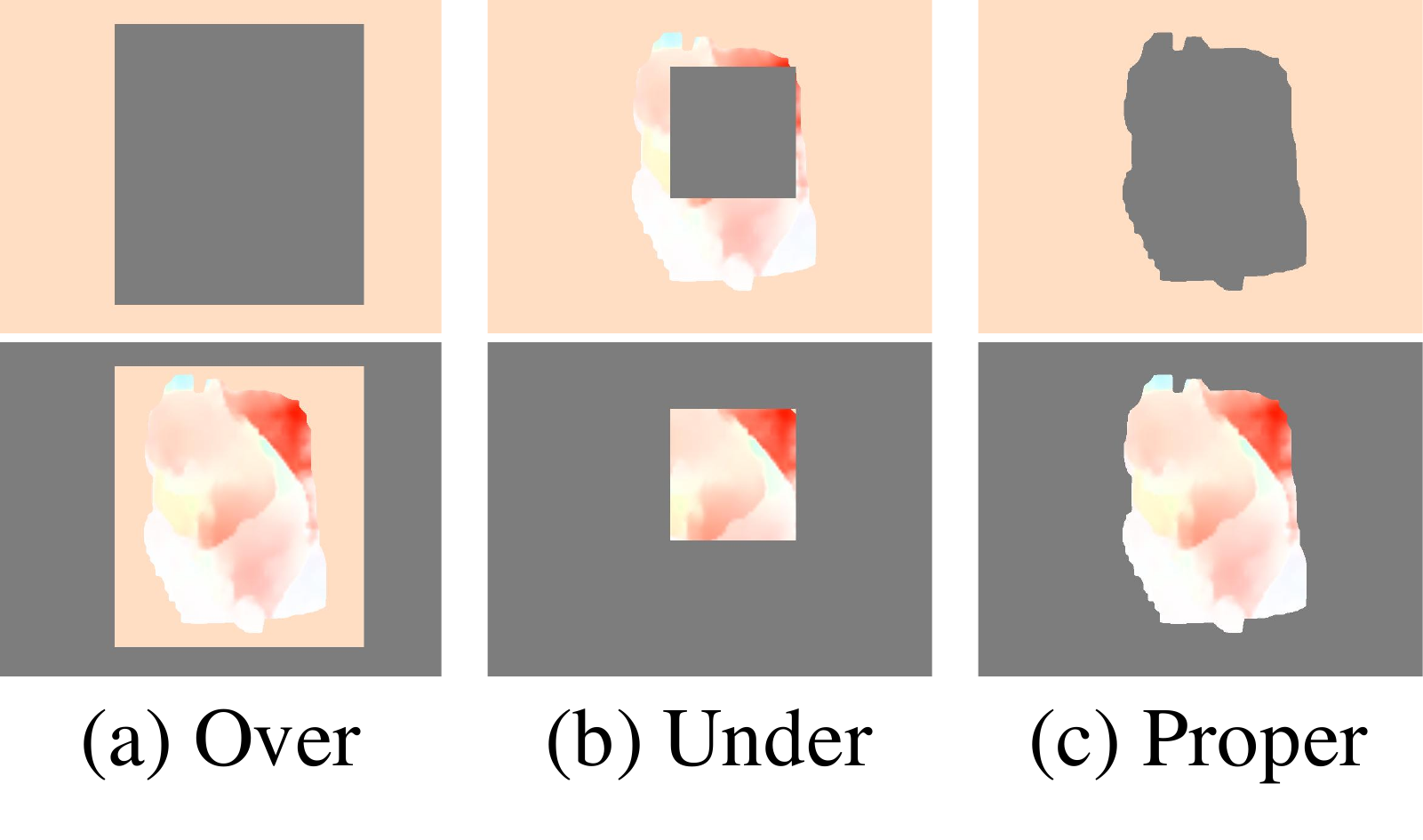}
		\vspace{-0.1cm}
		\caption{With image $\I$ in Fig.~\ref{fig:conditional}, only the mask in (c) minimizes mutual information between flow fields of the object (inside mask) and the background (outside mask).}
		\vspace{-0.2cm}
		\label{fig:over-under}
	\end{minipage}
\end{figure*}

\section{Related Work}

{\bf Motion segmentation} aims to identify independently moving regions in a video. 
Background subtraction assumes a static camera \cite{stauffer2000learning,elgammal2002background,radke2005image}, while scene dynamic models compensate for camera motion \cite{mittal2004motion,patwardhan2008robust,sheikh2009background}.
One could directly segment or cluster pixel-wise motion vectors  \cite{weiss1997smoothness,shi1998motion,kumar2008learning,papazoglou2013fast}, but this approach is prone to errors due to occlusions, singularities and discontinuities of the motion field.
To increase robustness, some employ pixel trajectories accumulated over multiple frames \cite{brox2010object,keuper2015motion,xu2015unsupervised,shen2018submodular} or patches, losing discriminative power. The hard trade-off between discriminability and robustness is a key challenge in unsupervised motion segmentation. Manual pixel-level annotations help set the trade-off, but in a non-scalable manner \cite{fragkiadaki2015learning,tokmakov2017learning,xiao2018monet,hu2018motion,dave2019towards}. Contextual Information Separation \cite{yang2019unsupervised} aims to bypass this trade-off without the need for human annotation, using a segmentation network to minimize the mutual information between the inside and the outside of putative motion regions. In the absence of temporal consistency, this procedure can be sensitive to motion errors  (Fig.~\ref{fig:effective-TC-SP}). Furthermore, \cite{yang2019unsupervised} cannot detect stationary objects as it relies on motion segmentation (Fig.~\ref{fig:motion-vs-static}).
Note, \cite{yang2020learning} applies \cite{yang2019unsupervised} with perceptual cycle-consistency to separating multiple objects, but they focus on learning object-centric representations.

{\bf Saliency prediction} aims to detect the most salient objects relative to their background or ``context.'' Saliency can be computed locally in a bottom-up fashion \cite{koch1987shifts,itti1998model,ma2003contrast} or globally \cite{zhai2006visual,achanta2009frequency,cheng2014global}, at multiple scales \cite{liu2006region,liu2010learning} and top-down \cite{tsotsos1995modeling}.
\cite{hou2007saliency} constructs the saliency map from the spectral residual and \cite{shen2012unified} performs low-rank matrix decomposition to detect salient objects. Separation can also be achieved by increasing the divergence of the feature distributions \cite{klein2011center}.
Despite advancements in the optimization for salient region segmentation \cite{comaniciu2002mean,felzenszwalb2004efficient,levinshtein2009turbopixels}, the quality of the predicted saliency depends highly on the features selected.
Currently, the best performing methods employ deep neural networks trained on labeled datasets \cite{zhang2017amulet,luo2017non,liu2018picanet,lin2014microsoft,li2014secrets}.
Recently, \cite{chen2019unsupervised,bielski2019emergence} proposed unsupervised adversarial salient object discovery models in single images.
Due to the variability of object appearance, these models are hard to train and the process is not scalable. On the other hand, \cite{zhang2018deep,nguyen2019deepusps} propose training deep networks using the pseudo-labels generated by conventional unsupervised methods. 
While an improvement over adversarial methods, performance still hinges on human prior knowledge through the selection of handcrafted features. 
We wish to avoid specifying features by instead articulating a criterion, namely that anything similar to what we have previously detected as objects should be salient. The model trained with this simple criterion outperforms the state of the art on unsupervised saliency prediction.

{\bf Video object segmentation} comprises a vast literature \cite{faisal2019exploiting,hu2018unsupervised,lao2018extending,yang2015self,wang2015saliency,wang2019ranet}; we focus on unsupervised methods that are more related to our work. It should be noted that the term ``unsupervised'' in video object segmentation only reflects absence of annotations at \textit{test time}, whereas we reserve the name for methods that involve no manual annotation during training, as well as testing. 
To segment moving objects, \cite{tokmakov2017learning} trains a network to directly output the segmentation from motion, which is then augmented by an appearance channel in \cite{jain2017fusionseg};  \cite{taylor2015causal} proposes a layered model for detachable objects using occlusion as primary cue. \cite{zhuo2019unsupervised} incorporates salient motion detection with object proposals;  \cite{koh2017primary} segments video into ``primary'' objects. 
A pyramid dilated bidirectional ConvLSTM is proposed in \cite{song2018pyramid} to extract spatial features at multiple scales, and \cite{lu2019see} introduces a global co-attention mechanism to capture scene context.
\cite{zhou2020motion} proposes an architecture that allows interaction between motion and appearance during the encoding process, while \cite{yang2019anchor,lai2019self} focuses on learning discriminative features for segmentation propagation and assumes the first frame been annotated.

\section{Method}
\label{sec:method}

An object that is detached from its surroundings induces an independent motion when projected onto a moving image \cite{ayvaci2011detachable}.
Here, we utilize this independence principle by minimizing the mutual information between the motions of an object and its context in Sec.~\ref{subsec:CIS_TC_SP}. 
In contrast to the Contextual Information Separation (CIS) criterion \cite{yang2019unsupervised}, we enforce that the separation is temporally consistent in a sequence, which directly improves \cite{yang2019unsupervised} by an average of 7\%. 
In Sec.~\ref{subsec:static_object}, we instantiate a static model that enables perception of stationary objects utilizing the detection of moving ones and the confidence measure computed from the mutual information. 
The interaction or mutual bootstrapping between motion segmentation and static perception is described in Sec.~\ref{subsec:joint_percept}. 
The overall method is illustrated in Fig.~\ref{fig:system}.

\subsection{Dynamic Model with Temporally Consistent Mutual Information Minimization}
\label{subsec:CIS_TC_SP}

Given an image $\I \in \mathbb{R}^{H\times W\times 3}$, the motion field $\uu$ defined on $\I$ is a random variable distributed according to $\p(\uu|\I)$, which is determined by a dataset of image sequences $\mathcal{D} = \{ \I^t_i \}_{i\leq N, t\leq T}$, where $N$ is the cardinality of the dataset and $T$ is the maximum number of images in a sequence. 
Particularly, for an instance $\iu \in \mathbb{R}^{H\times W\times 2}$, sampled from $\p(\uu|\I)$, there exists an image $\hat{\I}$ such that for any pixel $\x$ on $\I$, $\I(\x) = \hat{\I}(\x + \iu(\x))$ holds up to noise and occlusions, as shown in Fig.~\ref{fig:conditional}.

To detect objects that move independently in the scene, a motion segmentation network $\phi$ should generate masks $\m = \phi(\iu) \in \{0,1\}^{H\times W}$, such that the motions inside the mask $\m\odot\uu$ and outside the mask $(1-\m)\odot\uu$ are mutually independent conditioned on the image $\I$.
More explicitly, the conditional mutual information $\mi(\m\odot\uu, (1-\m)\odot\uu|\I)$ should be minimal.
Since the mutual information measures the difference between the Shannon entropy of the inside and its conditional entropy (conditioned) on the outside, simply minimizing the mutual information yields a trivial solution (empty set). 
One solution is to normalize the conditional mutual information by the entropy $\ent(\m\odot\uu|\I)$,
\begin{multline}
    \argmin_{\m} \dfrac{\mi(\m\odot\uu, (1-\m)\odot\uu|\I)}{\ent(\m\odot\uu|\I)}\\ 
    = \argmax_{\m} \dfrac{\ent(\m\odot\uu|(1-\m)\odot\uu,\I)}{\ent(\m\odot\uu|\I)}
\end{multline}
which is equivalent to maximizing the un-informativeness measured by the ratio on the right.
Again, this ratio can be maximized by simply setting $\m$ as the whole image domain, which results in over detection (Fig.~\ref{fig:over-under} (a)). 
Thus, if the detection is not accurate as in Fig.~\ref{fig:over-under} (c), 
the context will be rendered informative, vice versa.
Therefore it is necessary to add a symmetric term:
\begin{multline}
    \loss(\m; \I) = \dfrac{\ent(\m\odot\uu|(1-\m)\odot\uu,\I)}{\ent(\m\odot\uu|\I)} \\
    + \dfrac{\ent((1-\m)\odot\uu|\m\odot\uu,\I)}{\ent((1-\m)\odot\uu|\I)}
    \label{loss:symmetric-ratio}
\end{multline}
whose value is upper bounded by $2$, and can be maximized only when $\m$ accurately segments the object.

\begin{figure}[!t]
  \centering
  \includegraphics[width=.48\textwidth]{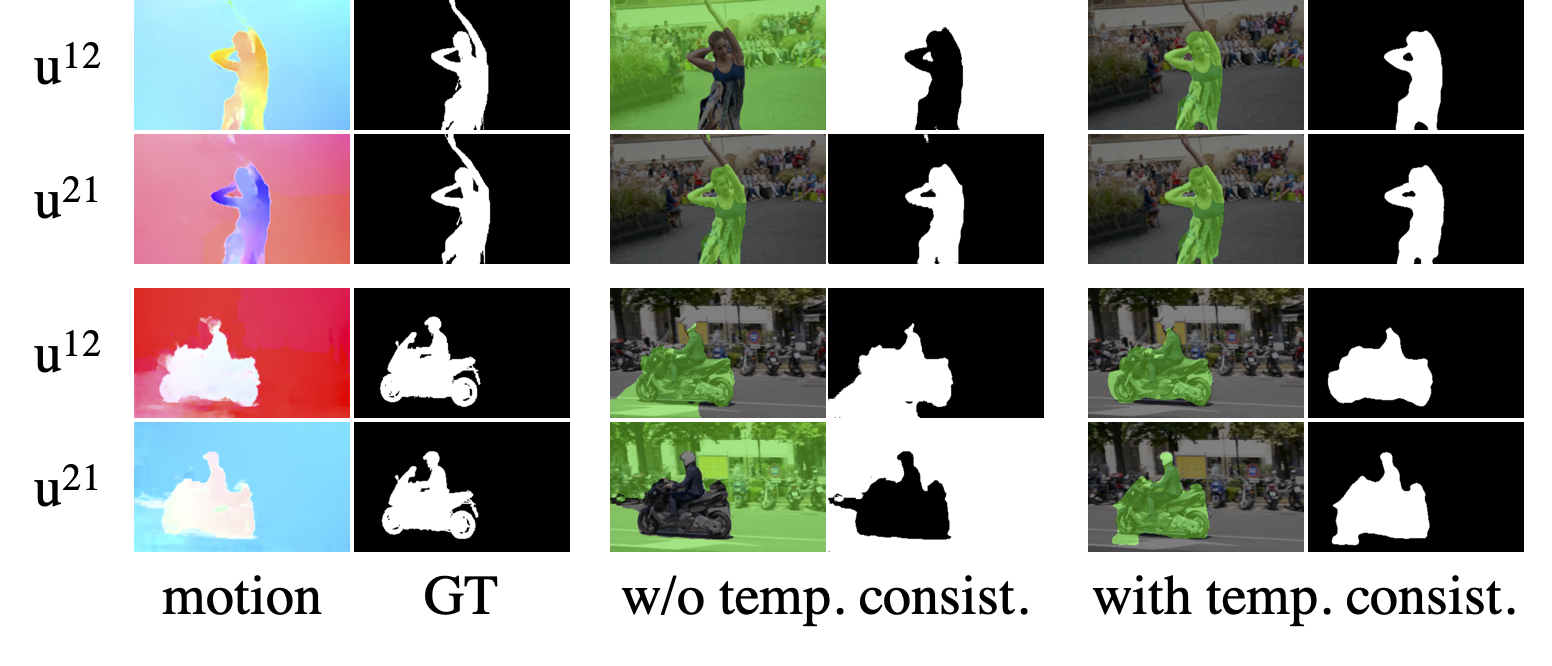}
  \vspace{-0.5cm}
  \caption{Temporally consistent mutual information minimization prevents label flipping (2nd and 3rd columns), and also improves segmentation accuracy (the motorbike).}
  \vspace{-0.2cm}
  \label{fig:effective-TC-SP}
\end{figure}

Similar to \cite{yang2019unsupervised}, we make the loss in Eq.~\eqref{loss:symmetric-ratio} computable by assuming Gaussian conditionals, and instantiating an adversarial inpainting network $\psi$ that computes the conditional means, e.g., $\psi(\m, (1-\m)\odot\iu, \I)$ estimates the conditional mean of $\m\odot\uu$ under $\p(\m\odot\uu|(1-\m)\odot\iu, \I)$. 
Both $\phi$ and $\psi$ can be trained adversarially: 
\begin{multline}
    \max_{\phi}\min_{\psi}\loss_{\A}(\phi,\psi; \I) = \\
    \dfrac{\sum_{\iu \sim \p(\uu|\I)} \| \m\odot\iu - \psi(\m, (1-\m)\odot\iu, \I) \|}
    {\sum_{\iu \sim \p(\uu|\I)}\| \m\odot\iu \| + \epsilon} \\
    + \dfrac{\sum_{\iu \sim \p(\uu|\I)} \| (1-\m)\odot\iu - \psi(1-\m, \m\odot\iu, \I) \|}
    {\sum_{\iu \sim \p(\uu|\I)} \| (1-\m)\odot\iu \| + \epsilon}
    \label{loss:CIS}
\end{multline}
with $\m = \phi(\iu)$, and $\| \cdot \|$ the $l^2$-norm. The constant $0<\epsilon\ll 1$ is to prevent numerical instability, and $\psi(\m, \emptyset, \I )$ is default to zeros.

Since Eq.~\eqref{loss:CIS} characterizes moving objects solely using motion, it is sensitive to variations in the motion field, resulting in label flipping and irregular segments due to failures of motion estimation shown in Fig.~\ref{fig:effective-TC-SP}. 
To resolve these issues, we introduce a temporal consistency constraint to reduce the instability in the motion segmentation model.

{\bf Temporal consistency.} Given two consecutive images $\I^1,\I^2$ from the same video sequence, 
we can compute the forward and backward optical flow $\iu^{12}, \iu^{21}$, and the predicted masks $\m^1 = \phi(\iu^{12}), \m^2 = \phi(\iu^{21})$. 
We would like the individually predicted masks to be temporally consistent,
in the sense that if we deform one onto the other using the flow fields, the two should look similar as they are the projections of the same object. Thus, we penalize the following warping difference to enforce temporal consistency:
\begin{multline}
    \loss_{\TC}(\phi; \iu^{12},\iu^{21}) = \sum_{\x \notin \mathbf{o}} | \m^1(\x) - \m^2(\x+\iu^{12}(\x)) | \\
    + | \m^2(\x) - \m^1(\x+\iu^{21}(\x)) |
    \label{loss:TC}
\end{multline}
with $\x$ the pixel index, and $\mathbf{o}$ the union of occlusions within the image domain, which can be easily estimated using the forward-backward identity criterion \cite{ince2008occlusion}. The reasoning is that inconsistencies of the predictions should only be penalized in the co-visible region.
Without introducing extra networks, Eq.~\eqref{loss:TC} effectively reduces instabilities in the motion segmentation compared to Eq.~\eqref{loss:CIS}, as shown in Fig.~\ref{fig:effective-TC-SP}.

{\bf Initial training for the dynamic model:} By enforcing temporal consistency, the dynamic model for independently moving object segmentation can be obtained through the following adversarial training:
\begin{multline}
    \max_{\phi}\min_{\psi}\loss_{\mathrm{\scriptscriptstyle D}}(\phi,\psi; \I^1,\iu^{12},\iu^{21}) \\
    = \loss_{\A}(\phi,\psi;\I^1) + \lambda_{\TC} \loss_{\TC}(\phi;\iu^{12},\iu^{21}) 
    \label{loss:cis-tc-sp}
\end{multline}
with $\lambda_{\TC}=-0.1$
($\lambda_{\TC}<0$ as $\phi$ maximizes $\loss_{\mathrm{\scriptscriptstyle D}}$). The effectiveness of optimizing Eq.~\eqref{loss:cis-tc-sp} is also numerically demonstrated in Tab.~\ref{tab:effective-TC-SP}.
Next, we describe the confidence-aware training for the static object model.

\begin{figure*}[!ht]
  \centering
  \includegraphics[width=0.85\textwidth]{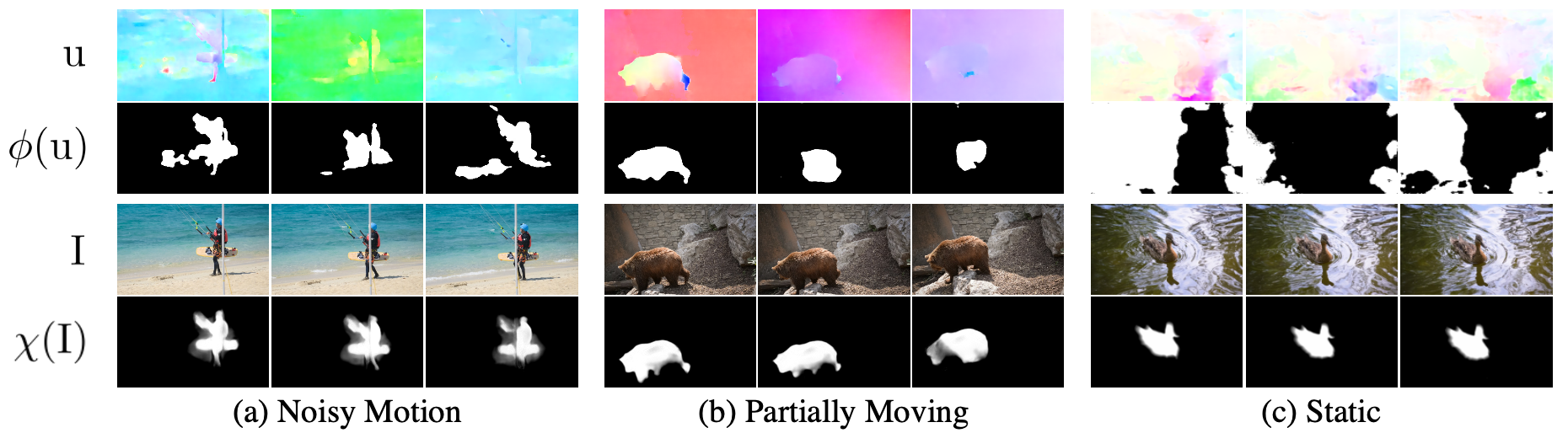}\
  \vspace{-0.0cm}
  \caption{Comparison between the dynamic and static object models. After the first round of learning from noisy motion segmentations ($\phi(\iu)$, second row) with the proposed confidence-aware adaptation, the static object model ($\chi(\I)$, fourth row) improves over the dynamic model on all cases where the motion is noisy (a), the object is partially static (b) or fully static (c).}
  \vspace{-0.2cm}
  \label{fig:motion-vs-static}
\end{figure*}

\subsection{Static Model with Confidence-Aware Update}
\vspace{-0.0cm}
\label{subsec:static_object}

In Sec.~\ref{subsec:CIS_TC_SP}, we describe a model that detects moving objects in a temporally coherent manner.
However, what if the objects stop moving or they have been moving in an indistinctive way? 
These present challenges for the dynamic model $\phi$, 
which relies on motion to signal the existence of an object (Fig.~\ref{fig:motion-vs-static}). 
Note in Fig.~\ref{fig:motion-vs-static}, the motion varies between frames, but the appearance of an object is temporally persistent, and when motion fades away, the image array still depicts the same object.

Our position is that, once a moving object is detected, we ought to be able to find it even in a still image. 
Thus, we propose to train a static object model to complement the dynamic model when there is no significant motion. 
We could directly train a segmentation network $\chi$ to predict objects from a single image, utilizing the output of $\phi$ as the pseudo labels, by maximizing the F-measure \cite{nguyen2019deepusps} commonly used for salient object detection:
\begin{equation}
    F_{\alpha}(\chi(\I), \overline{\m}) = (1+\alpha^2)\dfrac{\rho(\chi(\I),\overline{\m})\gamma(\chi(\I),\overline{\m})}{\alpha^2\rho(\chi(\I),\overline{\m})+\gamma(\chi(\I),\overline{\m})}
    \label{loss:f-beta}
\end{equation}
with $\rho, \gamma$ the precision and recall between the prediction $\chi(\I)$ and the motion mask $\overline{\m}$ generated by $\phi$. And $\alpha^2$ is default to 1.5 if not explicitly mentioned.

\begin{figure*}[!t]
	\centering
	\begin{minipage}[t]{.45\textwidth}
	\centering
    \includegraphics[width=.8\textwidth]{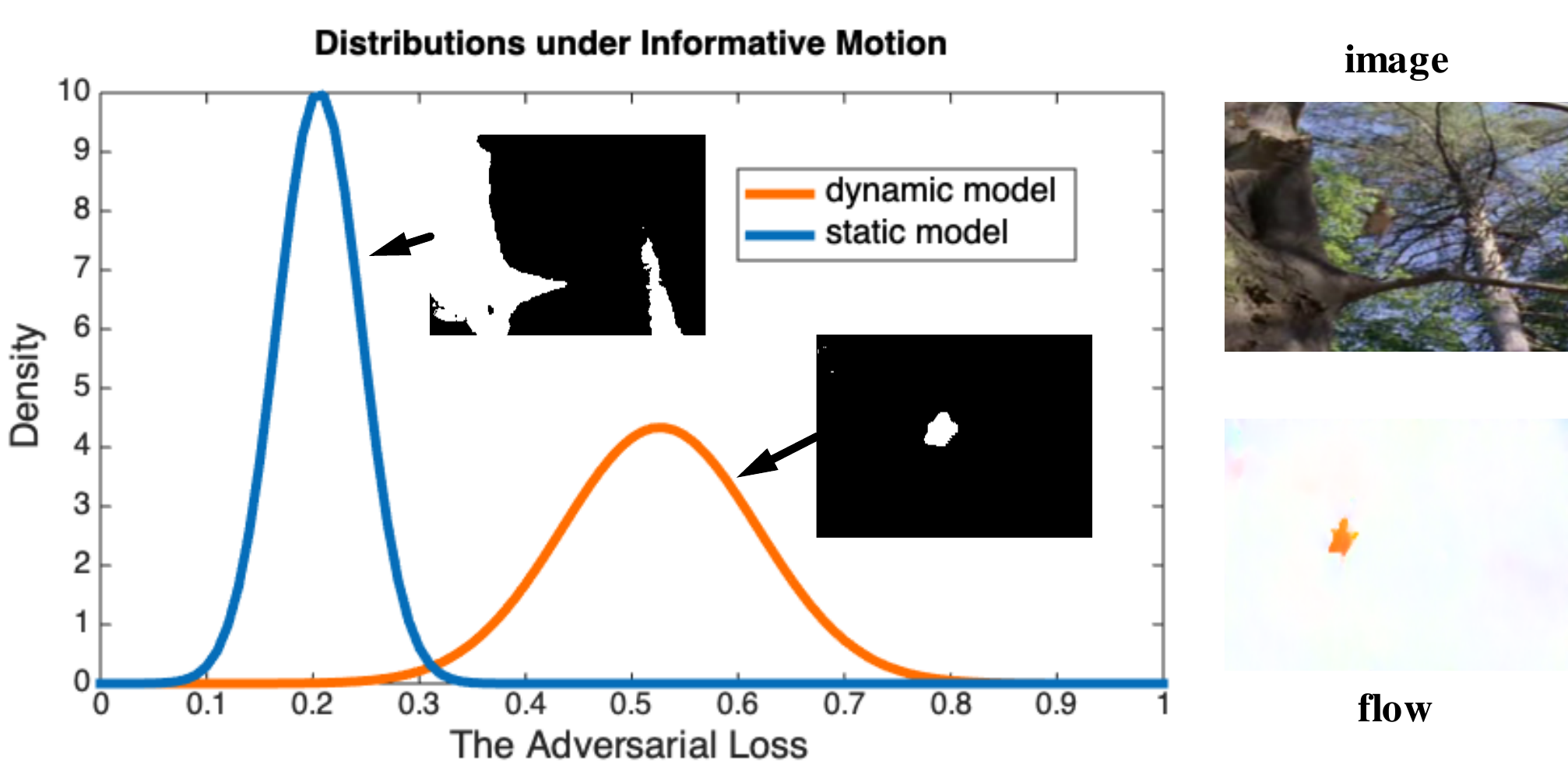}
    \vspace{-0.0cm}
    \caption{Distribution of $\loss_{\A}$ computed using the output of the dynamic model $\phi$ (orange) and the static model $\chi$ (blue) on the birdfall sequence, which shows that $\loss_{\A}$ is a good indicator of reliable masks when the motion field is informative, as verified by the gap between the distributions of $\loss_{\A}$.}
    \vspace{-0.3cm}
    \label{fig:birdfall}
	\end{minipage}
	\hspace{1.0cm}
	\begin{minipage}[t]{.45\textwidth}
	\centering
    \includegraphics[width=.8\textwidth]{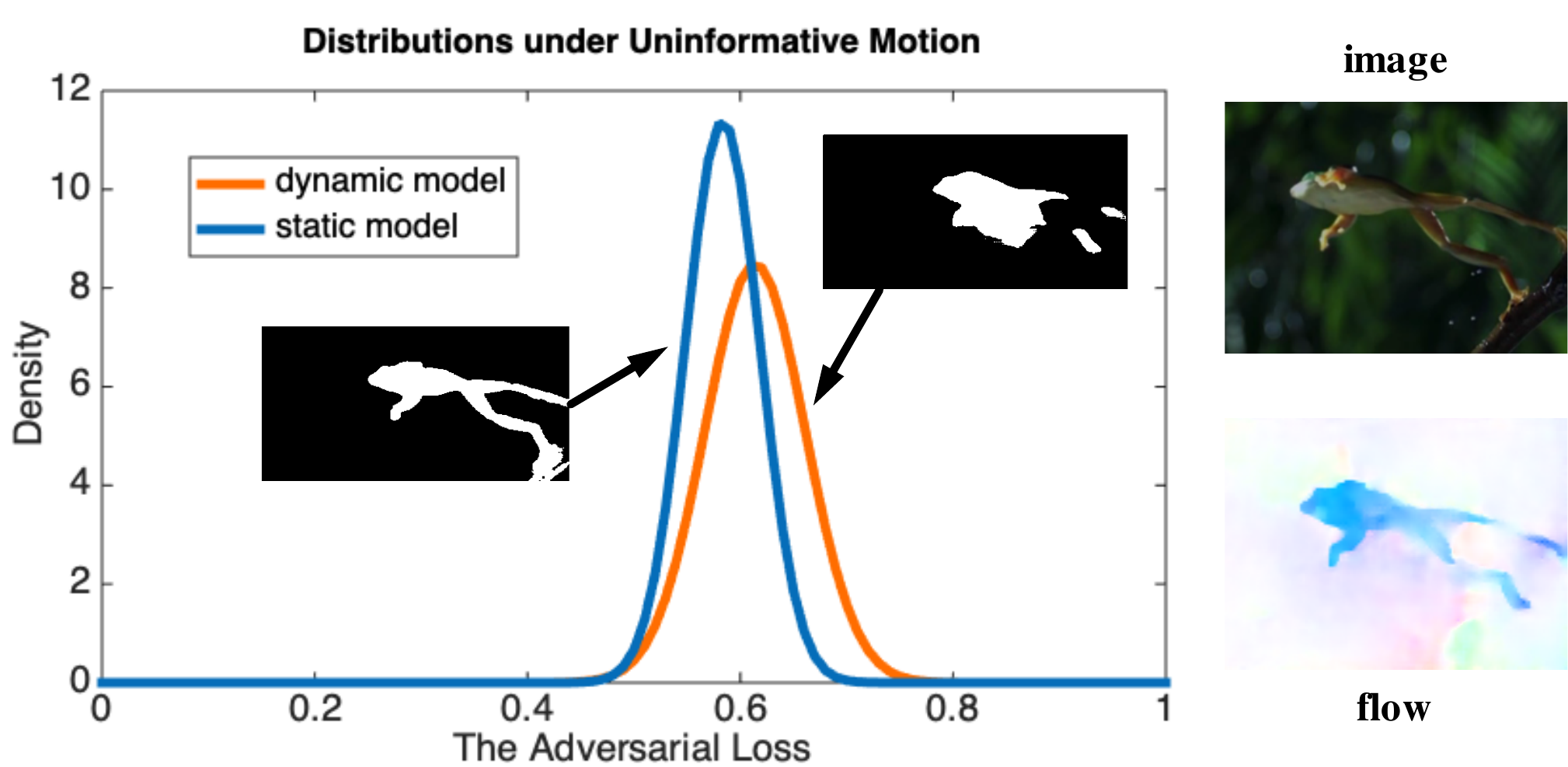}
    \vspace{-0.0cm}
    \caption{Distribution of $\loss_{\A}$ for the dynamic model $\phi$ (orange) and the static model $\chi$ (blue) on the frog sequence, which shows that when the motion is not informative or erroneous, the static model can work better, thus the difference in the values of $\loss_{\A}$ is small.}
    \vspace{-0.3cm}
    \label{fig:frog}
	\end{minipage}
\end{figure*}

However, directly learning from all motion masks is counter-productive, as these are quite noisy especially when motion is uninformative (Fig.~\ref{fig:motion-vs-static}). 
To address this issue, we propose to use the loss $\loss_{\A}$ (Eq.~\eqref{loss:CIS}) as a confidence measure on the reliability of the motion masks: 
{\it if the motion is uninformative, 
the reconstruction from the context will be accurate, 
thus $\loss_{\A}$ will be small and vice versa, if the motion is distinctive, $\loss_{\A}$ will be large due to a bad reconstruction} (See Fig.~\ref{fig:birdfall} and Fig.~\ref{fig:frog} for a demonstration.)
We propose using the difference between the values of $\loss_{\A}$ to perform a confidence-aware adaptation via the following:
\begin{multline}
    \loss_{\mathrm{\scriptscriptstyle \chi}}(\chi; \I,\overline{\m},\chi') = \lambda_{F}F_{\alpha}(\chi(\I),\chi'(\I))\\
    + \max(\loss_{\A}(\overline{\m})-\loss_{\A}(\chi'(\I))-\delta,0)F_{\alpha}(\chi(\I),\overline{\m})
    \label{loss:static-seg}
\end{multline}
Note that the pseudo-label $\overline{\m}$ is only effective when it has a larger $\loss_{\A}$ than the one predicted by $\chi'$, 
which is a copy of an earlier $\chi$. 
In other words, if the output of the dynamic model is not confident enough, $\chi$ retains its own prediction, and the first term (moving average) is to ensure that $\chi$ is updated smoothly.
We set $\loss_{\A}(\chi'(\I))=0$ and $\lambda_F=0$ the first time $\chi$ is trained, then $\lambda_F=1.0$.

When updated, $\chi$ learns a model of objects based on their appearance, so we would expect $\chi$ to detect stationary objects which have been seen moving before. 
Indeed, we find that $\chi$ is able to detect static object that has never been observed moving as shown in Fig.~\ref{fig:motion-vs-static} (3rd column), which confirms that a general concept of objects can be learned through the observations of moving ones. Next we detail the proposed dynamic-static bootstrapping scheme for a continuous learning of objects.

\subsection{Dynamic-Static Bootstrapping}
\vspace{-0.0cm}
\label{subsec:joint_percept}

Once the static object model is learned by $\chi$, it can be used to modulate the detection in general scenes, even where the motion of the objects is unknown.
It is also possible that the predictions of $\phi$, which employs motion information are imperfect. 
The output of $\chi$ can then provide complementary information to strengthen the dynamic model. 
We feed $\chi$ back into the training of the dynamic model as an experiential prior of objectness based on the photometric information. 
The dynamic model reinforced by the objectness prior is:
\begin{multline}
    \max_{\phi}\min_{\psi} \loss_{\J}(\phi,\psi; \I^1,\iu^{12},\iu^{21},\chi) =\\ \loss_{\mathrm{\scriptscriptstyle D}}(\phi,\psi; \I^1,\iu^{12},\iu^{21}) + \lambda_{\mathrm{obj}} F_{\alpha}(\phi(\iu^{12}),\chi(\I^1))
    \label{loss:joint-training}
\end{multline}
with $\loss_{\mathrm{\scriptscriptstyle D}}$ the adversarial motion segmentation loss in Eq.~\eqref{loss:cis-tc-sp}, and the second term measures the similarity between the motion mask and the static object prior ($\lambda_{\mathrm{obj}}=1.0$).
Besides learning from motion information to detect moving objects, $\phi$ is now able to leverage photometric cues that facilitate the detection under circumstances where objects become stationary or move extremely slowly. 

Moreover, improved dynamic model could yield better pseudo-labels that help training a more accurate static object model (Eq.~\eqref{loss:static-seg}), which can then be used to facilitate the learning of the former in a synergistic loop (Eq.~\eqref{loss:joint-training}), thus the name ``Dynamic-Static Bootstrapping.''
The overall training procedure is presented in Algorithm~\ref{algo:all}.

\begin{algorithm}[!h]
    \KwResult{$\phi$: dynamic model; $\chi$: static object model; $\psi$: conditional inpainting network}
    Initialize $\phi,\psi$ by optimizing $\loss_{\mathrm{\scriptscriptstyle D}}$ (Eq.~\eqref{loss:cis-tc-sp}), set k=0\;
    \While{k<3}{
    k = k+1\;
    Update the static model $\chi$ using $\loss_{\mathrm{\scriptscriptstyle \chi}}$ (Eq.~\eqref{loss:static-seg})\;
    Update the dynamic model $\phi$ using $\loss_{\J}$ (Eq.~\eqref{loss:joint-training})\;
    }
    \caption{Dynamic-Static Bootstrapping}
    \vspace{-0.1cm}
    \label{algo:all}
\end{algorithm}

\section{Implementation}
\label{sec:implement}

\textbf{Dynamic model $\phi$} uses the Deeplab architecture \cite{chen2017deeplab}, with the initialization weights as in \cite{zhang2018deep,nguyen2019deepusps}. $\phi$ takes as input the estimated flow between two randomly chosen frames from the same video with the maximum interval equals to three. Optical flow is produced by PWCNet \cite{sun2018pwc}, which is trained on synthetic data.
The output of $\phi$ is a two channel softmax score.
In total, $\phi$ has 23M trainable parameters and can run in 19 fps during inference.

\textbf{Static model $\chi$} uses similar architecture as the dynamic model, but takes a single RGB image as input. The output is also a two channel softmax score. The total number of parameters is 21 M, and $\chi$ can run at 22 fps at inference.

{\bf Training details.} For the initial training of the dynamic model $\phi$, we alternate between updating $\phi$ for three steps and updating $\psi$ for one step, up to 30 epochs on the training set of each dataset, using an Adam optimizer with lr=1e-4, beta1=0.9, and beta2=0.999. 
The static model is then trained up to 15 epochs, using an Adam optimizer with lr=2e-5, beta1=0.9, and beta2=0.999.
As described in Algorithm~\ref{algo:all}, the dynamic-static bootstrapping runs for three iterations. On average, each iteration takes six hours to converge, and the whole training procedure can be finished within a day.

\section{Experiments}
\label{sec:exp}

{\bf Datasets:} For {\it video object segmentation}, we train and test our model on three commonly used video object segmentation datasets: DAVIS \cite{perazzi2016benchmark}, FBMS \cite{ochs2013segmentation}, and SegTrackV2 \cite{li2013video} (Tab.~\ref{tab:vos-benchmarks}).
We also test our trained model on two other datasets, DAVIS17 \cite{pont20172017} and Youtube-VOS \cite{xu2018youtube}, to check how well our method generalizes (Tab.~\ref{tab:D17-YVOS}).

DAVIS consists of high-resolution videos (30 for training and 20 for validation) depicting the primary object moving in the scene with pixel-wise annotations for each frame. 
FBMS contains videos of multiple moving objects, providing test cases for multiple object segmentation. FBMS has sparsely annotated 59 video sequences, with 30 sequences for validation.
SegTrackV2 contains 14 densely annotated videos. 
These videos constitute the only source of training data for our unsupervised motion perception module $\phi$.
Youtube-VOS contains 4,453 videos and 94 object categories, and DAVIS17 consists of 150 videos.

To evaluate $\chi$ on static object segmentation, we test on three major saliency prediction datasets: MSRA-B \cite{jiang2013salient} (5000 images), ECCSD \cite{shi2015hierarchical} (1000 images) and DUT \cite{yang2013saliency} (5168 images). All three datasets are annotated with pixel-wise labels for each image. 
These saliency datasets contain objects from a much broader span of categories, such as road signs, statues, flowers, etc., that are never seen moving in the training videos 
We evaluate the static object model learned from only video objects 
on these saliency benchmarks, 
to check its transferability to different instances from seen categories and unseen categories.

\subsection{Effectiveness of Temporally Consistent Mutual Information Minimization}

To verify the effectiveness of the temporally consistent mutual information minimization proposed in Sec.~\ref{subsec:CIS_TC_SP} for bottom-up motion segmentation, we compare to the baseline CIS \cite{yang2019unsupervised} that trains a segmentation network using only Eq.~\eqref{loss:CIS}.
We train both CIS \cite{yang2019unsupervised} and our model described in Eq.~\eqref{loss:cis-tc-sp} on the unlabeled videos from DAVIS, and then test on the validation set of DAVIS. We also report the scores by directly applying the model trained on DAVIS to FBMS and SegTrackV2 in Tab.~\ref{tab:effective-TC-SP} to check the generalization on different domains. 
The performance is measured by mean-Intersection-over-Union (mIoU), and the relative weights used in our model are $\lambda_{\TC}=1.0$. As shown in Tab.~\ref{tab:effective-TC-SP}, our dynamic model (Eq.~\eqref{loss:cis-tc-sp}) consistently outperforms CIS (Eq.~\eqref{loss:CIS}) on all three video object segmentation benchmarks by 7\%, which confirms that temporal consistency is a critical component in our dynamic model.

\begin{table}[!h]
	\centering
    \begin{tabular}{cccc}
    \toprule
    & {DAVIS} & {FBMS} & {SegTV2}\\
    \midrule
    CIS \cite{yang2019unsupervised} & 59.2 & 36.8 & 45.6 \\
    Ours & {\bf 62.4} & {\bf 40.0} & {\bf 49.1} \\
    \bottomrule
    \end{tabular}
    \vspace{0.05cm}
    \caption{Our temporally consistent dynamic model v.s. CIS \cite{yang2019unsupervised}.}
    \label{tab:effective-TC-SP}
\end{table}

\subsection{Effectiveness of Confidence-Aware Adaptation}

Here we check the effectiveness of the confidence-aware adaptation scheme proposed in Sec.~\ref{subsec:static_object} in precluding counter-productive self-learning.
We first train a motion model $\phi$ on the training data from DAVIS. 
Then we train two static models $\chi$: $\chi(F_{\alpha})$ using Eq.~\eqref{loss:f-beta} (not confidence-aware), the other $\chi(\loss_{\chi})$ using Eq.~\eqref{loss:static-seg} (confidence-aware).
We set $\delta$ in Eq.~\eqref{loss:static-seg} to 0.2 and $\lambda_{F}$ to 1.0, which are fixed for the future experiments.
We compare the performance of the static models on the DAVIS validation set using mIoU. Further, we perform the same evaluation on both FBMS and SegTrackV2, and report the scores in Tab.~\ref{tab:adaptive-bootstrap}. As shown, with the confidence-aware adaptive bootstrapping loss Eq.~\eqref{loss:static-seg}, the static object model $\chi$ consistently improves over its counter-part on the three benchmarks, confirming the importance of uncertainty estimation in self-supervised learning.

\begin{table}[!h]
\centering
    \begin{tabular}{cccc}
    \toprule
    & {DAVIS} & {FBMS} & {SegTV2} \\
    \midrule
    $\chi(F_{\alpha})$   &  73.8  & 65.5 & 65.7  \\
    $\chi(\loss_{\chi})$ & {\bf 78.2} & {\bf 68.7} & {\bf 69.3} \\
    \bottomrule
    \end{tabular}
    \vspace{0.1cm}
    \caption{Static model $\chi(F_{\alpha})$ vs. $\chi(\loss_{\chi})$. The latter is trained with confidence-aware adaptation.}
    \label{tab:adaptive-bootstrap}
\end{table}

\begin{figure}[!t]
\centering
  \includegraphics[width=0.47\textwidth]{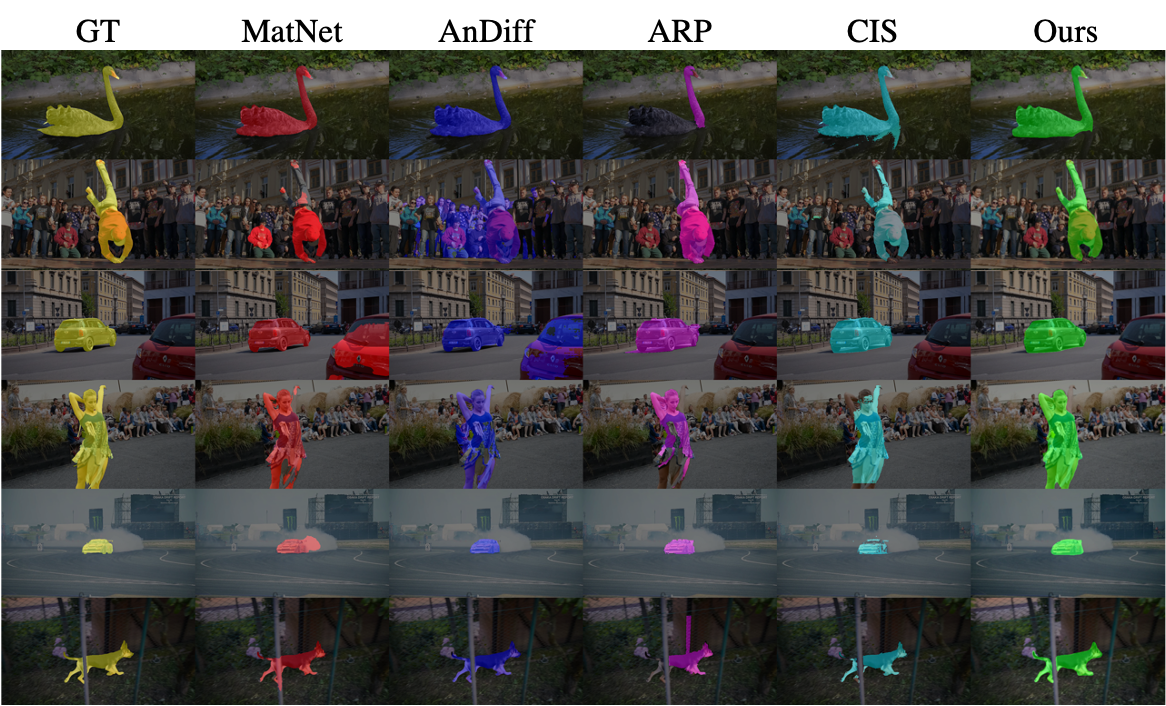}
  \caption{Qualitative comparison to the top two methods from each category on the DAVIS benchmark. MatNet \cite{zhou2020motion}, AnDiff \cite{yang2019anchor} (supervised) and ARP \cite{koh2017primary}, CIS \cite{yang2019unsupervised} (unsupervised).}
  \vspace{-0.3cm}
  \label{fig:vis-davis}
\end{figure}

\begin{table*}[!h]
\setlength{\tabcolsep}{6pt}
  \centering
  \begin{tabular}{lcccclccc}
    \toprule
    \multicolumn{5}{c}{Supervised Methods} & \multicolumn{4}{c}{Unsupervised Methods} \\
    \midrule
     & {\small $\#$ Annot.} & {\small DAVIS} & {\small FBMS(J/F)} & {\small SegTV2} & & {\small DAVIS} & {\small FBMS} & {\small SegTV2} \\
     \midrule
     {\small MATNet \cite{zhou2020motion}}  & 14,000 & 82.4 & {\bf 76.1} / ----- & 50.2 & {\small ARP \cite{koh2017primary}} & 76.2 & 59.8 & 57.2 \\
     {\small AnDiff \cite{yang2019anchor}}  & 2,000 & 81.7 & ----- / 81.2 & 48.3 &
     {\small ELM \cite{lao2018extending}} & 61.8 & 61.6 & -- \\
     {\small COSNet \cite{lu2019see}}  & 17,000 & 80.5 & 75.6 / ----- & 49.7 &
     {\small FST \cite{papazoglou2013fast}} & 55.8 & 47.7 & 47.8 \\
     {\small EPONet \cite{faisal2019exploiting}}  & 2,000+ & 80.6 & -- & { 70.9} & 
     {\small NLC \cite{faktor2014video}} & 55.1 & 51.5 & 67.2 \\
     {\small PDB \cite{song2018pyramid}}  & 17,000 & 77.2 & 74.0 / 81.5 & 60.9 & 
     {\small SAGE \cite{wang2015saliency}} & 42.6 & 61.2 & 57.6 \\
     {\small LVO \cite{tokmakov2017learning}}  & 2,000+ & 75.9 & 65.1 / 77.8 & 57.3 & 
     {\small STP \cite{hu2018unsupervised}} & 77.6 & 60.8 & 70.1 \\
     {\small FSEG \cite{jain2017fusionseg}} & 10,500 & 70.7 & 68.4 / ----- & 61.4 & 
     {\small CIS \cite{yang2019unsupervised}} & 71.5 & 63.6 & 62.0 \\
     {\small Ours } & 2,000 & {\bf 82.8} & 75.8 / {\bf 82.0} & {\bf 74.2} & Ours & {\bf 80.0} & {\bf 73.2} & {\bf 74.2}\\
    \bottomrule
  \end{tabular}
\vspace{0.1cm}
  \caption{Quantitative comparison on video object segmentation benchmarks with both supervised and fully unsupervised methods.}
  \vspace{-0.1cm}
  \label{tab:vos-benchmarks}
\end{table*}

\begin{table*}
\setlength{\tabcolsep}{4pt}
\centering
    \begin{tabular}{cccccccccccc}
    \toprule
    \multicolumn{5}{c}{Supervised Methods} & \multicolumn{4}{c}{Unsupervised Methods} \\
    \midrule
      & {\small \it DSS \cite{hou2017deeply}} & {\small \it NDF \cite{luo2017non}} & {\small \it SR \cite{wang2017stagewise}} & {\small RBD \cite{zhu2014saliency}} & {\small DSR \cite{li2013saliency}} & {\small HS \cite{zou2015harf}} & {\small CHS \cite{wangiccv17}} & {\small SBF \cite{yan2013hierarchical}}
     & {\small USD \cite{zhang2018deep}} & {\small DUSPS \cite{nguyen2019deepusps}} & {\small Ours}  \\ 
    \midrule
    {\small ECCSD}  & 87.9 & 89.1 & 82.6 & 65.2 & 63.9 & 62.3 & 68.2 & 78.7 & 87.8 & 87.4 & \textbf{88.1}  \\
    {\small MSRA-B}  & 89.4 & 89.7 & 85.1 & 75.1 & 72.3 & 71.3 & 79.8 & -- & 87.7 & \textbf{90.3}  & 89.7  \\
    {\small DUT}    & 72.9 & 73.6 & 67.2 & 51.0 & 55.8 & 52.1 & -- & 58.3 & 71.6 & 73.6 & \textbf{73.9} \\
    \bottomrule
    \end{tabular}
    \vspace{0.1cm}
    \caption{Quantitative results on saliency prediction (or salient object detection) benchmarks.}
    \vspace{-0.1cm}
    \label{tab:saliency-benchmarks}
\end{table*}

\subsection{Improvement from Bootstrapping}

In Tab.~\ref{tab:multiple-rounds}, 
we show the improvement after multiple rounds of bootstrapping for each model as described in Algorithm~\ref{algo:all} (dynamic-static bootstrapping).
The reported score is the mIoU of the segmentation on the three video object segmentation benchmarks.
This demonstrates the effectiveness of training in a synergistic loop with our dynamic-static bootstrapping strategy.

\begin{table}[!h]
	\centering
	\begin{tabular}{cccc}
    \toprule
    \# of Rounds   & DAVIS & FBMS & SegTV2 \\
    \midrule
    1 &  73.8  & 65.5 & 65.7  \\
    2 & 79.2 & 71.7 & 73.1 \\
    3 & {\bf 80.0} & {\bf 73.2} & {\bf 74.2} \\
    \bottomrule
    \end{tabular}
    \vspace{0.1cm}
    \caption{The mIoU improves over multiple rounds of dynamic-static bootstrapping.}
    \label{tab:multiple-rounds}	
\end{table}

\subsection{Static Model Improves with the Number of Training Videos}

One characteristic of our method is that the static model $\chi$ improves over time as more and more objects are seen moving through the bottom-up motion detection module $\phi$. 
To verify, we construct a collection of videos $S$ by combining the three aforementioned video object segmentation datasets (in total there are 123 video sequences). 
We randomly partition them into 10 subsets $\{s_k\}_{k=1}^{10}$, each contains around 12 video sequences.
Correspondingly, we train 5 static object models $\{\chi_i\}_{i=0}^{4}$ by performing Algorithm~\ref{algo:all}. 
The training set for each $\chi_i$ is $\{s_k\}_{k=1}^{2i+1}$, such that $\chi_i$ with a larger $i$ is exposed to more video sequences. 
Each $\chi$ is evaluated on the union of the three saliency datasets mentioned above (in total 11,000 testing images).
In Tab.~\ref{tab:chi-improves}, 
we report the performance of $\chi_i$'s, measured in terms of mIoU (with standard deviation computed across five runs).
As shown in Tab.~\ref{tab:chi-improves}, 
when the number of the observed videos increases, the performance of the static object model also improves, which is consistently observed across multiple runs.

\begin{table}[!h]
\centering
    \begin{tabular}{cccccc}
    \toprule
    $\#$ of videos & 12 & 36 & 60 & 84 & 108 \\
    \midrule
    mIoU & 48.3 & 52.4 & 54.9 & 58.8 & 61.4 \\
    Std. Dev. & 1.99 & 2.37 & 2.30 & 1.17 & 1.56 \\
    \bottomrule
    \end{tabular}
    \vspace{0.1cm}
    \caption{Static model $\chi$ improves over time as more videos are observed.}
    \label{tab:chi-improves}	
\end{table}

\begin{figure}[!t]
   \centering
   \includegraphics[width=0.47\textwidth]{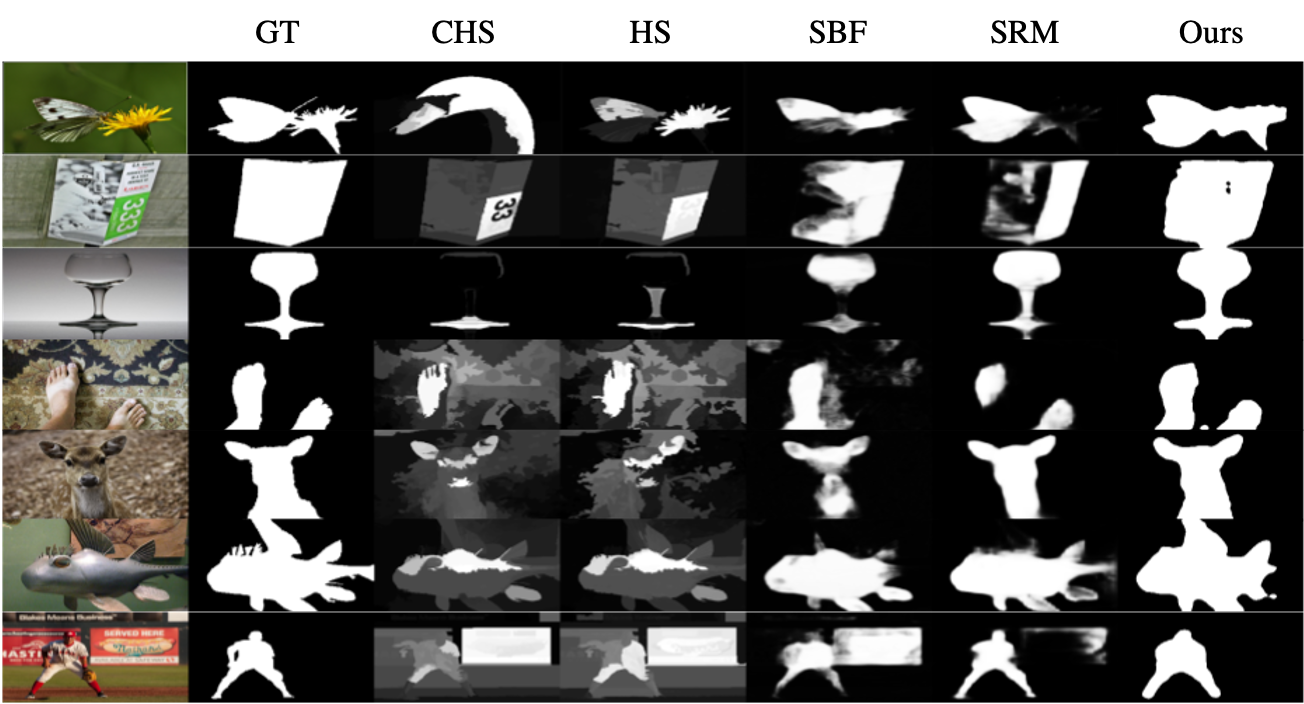}
   \vspace{-0.1cm}
   \caption{Visual results on the saliency prediction benchmarks. CHS \cite{yan2013hierarchical}, HS \cite{zou2015harf} (unsupervised), SBF \cite{wangiccv17} (unsupervised learning-based), SR \cite{wang2017stagewise} (supervised).}
   \vspace{-0.1cm}
   \label{fig:failure-case}
\end{figure}

\subsection{Video Object Segmentation Benchmarks}

To check the effectiveness of the proposed dynamic-static bootstrapping in learning better motion segmentation, 
we evaluate on the benchmarks for video object segmentation.
Since video object segmentation focuses on moving objects, we weigh the predictions from the dynamic model with the predictions from the static model to emphasize the detection of moving ones.
Similar to \cite{yang2019unsupervised}, CRF postprocessing is performed to get our final results. 
We compare to top-performing unsupervised (no annotations are involved) and supervised (annotations are involved during the training phase) methods. 
The performance is measured by mIoU. 
As shown in Tab.~\ref{tab:vos-benchmarks}, our method achieves the top performance on all three video object segmentation benchmarks among fully unsupervised methods. 
To compare with methods that utilize manual annotations (Supervised), we finetune our model on the DAVIS training set with 2000 annotations.
We have also listed the number of annotations used by other supervised methods in Tab.~\ref{tab:vos-benchmarks}.
Again, our model achieves the top performance using the least amount of manual annotations among all the supervised methods.
In Tab.~\ref{tab:D17-YVOS} we show the results by testing our trained models on two other video object segmentation benchmarks DAVIS17 and Youtube-VOS. Our model still achieves competitive performance compared to the state of the art and demonstrates good generalization.

\subsection{Unsupervised Salient Object Detection}

In Tab.~\ref{tab:saliency-benchmarks}, we evaluate the learned object prior (static object model) through the task of salient object detection in images. Note that the top-performing methods on unsupervised salient object detection all rely on handcrafted methods either as the primary procedure, or as a subprocess. 
Among all top-performing ones, we are, to the best of our knowledge, the only one that does not rely on any handcrafted features.
We refine the static model $\chi$ by performing CRF on its predictions, and by one round of self-training with the CRF refined masks \cite{nguyen2019deepusps}. 
By leveraging the object prior learned through videos, we can approach and surpass the state of the art. Even when compared with top-performing supervised methods ({\it DSS, NDF, SR} in Tab.~\ref{tab:saliency-benchmarks}), our method still achieves competitive performance with no explicit annotation.

\begin{table}
\setlength{\tabcolsep}{2pt}
\centering
    \begin{tabular}{lccccc}
    \toprule
    &{\small MATNet \cite{zhou2020motion}}& {\small Andiff \cite{yang2019anchor} }& {\small ARP \cite{koh2017primary} }& {\small CIS \cite{yang2019unsupervised}} & {\small Ours}\\
    \midrule
    {\small DAVIS17} & 58.6 & 57.8 & 50.2 & 53.1 & {\bf 58.9} \\
    {\small YTVOS}  & -- & 46.1 & 28.7 & 15.6 & {\bf 47.2}\\
    \bottomrule
    \end{tabular}
    \vspace{0.1cm}
    \caption{Results on DAVIS17 \cite{pont20172017} and Youtube-VOS \cite{xu2018youtube} datasets.}
    \vspace{-0.1cm}
    \label{tab:D17-YVOS}	
\end{table}

\section{Discussion}

We have presented a method to learn how to segment objects in images that exploits temporal consistency in their motion, observed in training videos, to bootstrap a top-down model. 
The definition of what constitutes an object is implicit in the method and in the datasets used for training. 
This may appear to be a limitation, as training on different datasets may yield different outcomes. 
However, what constitutes an object, or even a segment of an image, is ultimately not objective: In Fig.~\ref{fig:effective-TC-SP}, is the object a person? Or the motorcycle they are riding? The union of the two? The helmet they are wearing? All of the above? We let the evidence bootstrap the definition: If the motion at the resolution of the first video shows the human and bike moving as a whole, we do not know any better than to consider them one object. 
If, in later video, a human is seen without a motorcycle, they will be an independent object thereafter.  
Admittedly, our model does not capture the fact that a proper object model should segment instances and enable multiple memberships for each point: A pixel on the helmet is part of the object, but also of the person, and the rider, and so on. 
We also do not exploit side information from other modalities. 
Nonetheless, despite the complete absence of annotation requirements, our method edges out methods that exploit manual annotation, so we believe it to be a useful starting point for further development of more complete object segmentation methods.

\subsubsection*{Acknowledgment}

Research supported by ARO W911NF-17-1-0304 and ONR N00014-17-1-2072.

\section{Appendix}

\subsection{From Eq.~\ref{loss:symmetric-ratio} to Eq.~\ref{loss:CIS}}

Under the Gaussian assumption, the Shannon entropy is proportional to the variance of the Gaussian. So Eq.~\ref{loss:CIS} can be obtained by replacing Shannon entropies in Eq.~\ref{loss:symmetric-ratio} with the estimated variances.

\begin{figure*}[!t]
  \centering
  \includegraphics[width=0.99\textwidth]{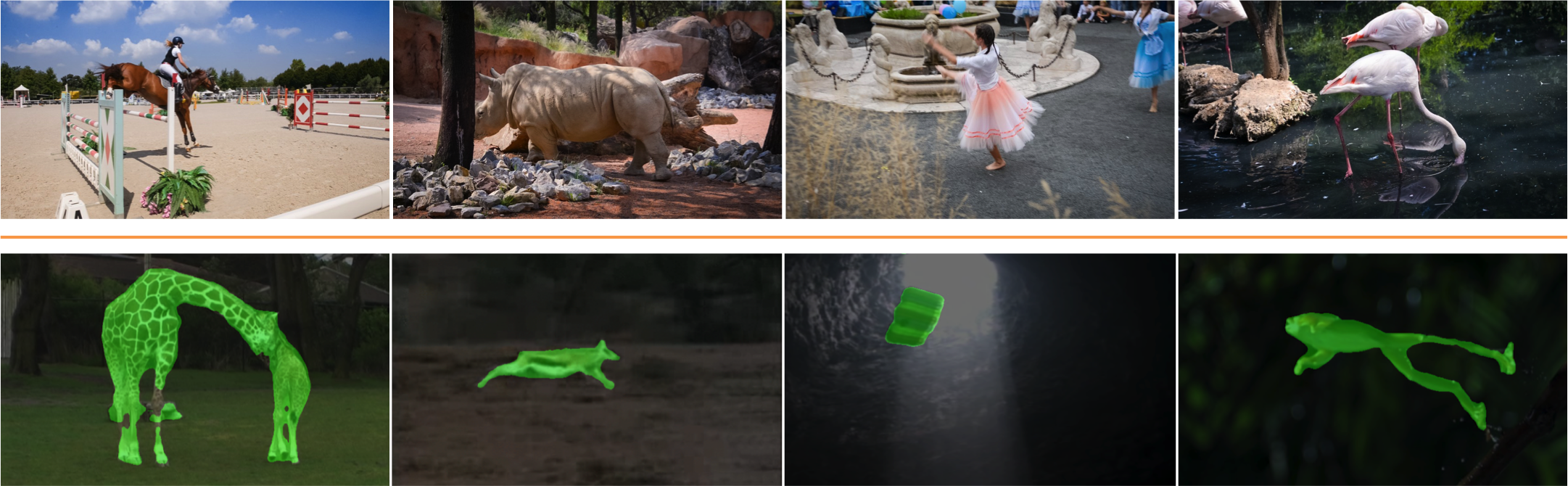}
  \caption{Top: frames from sequences in DAVIS \cite{perazzi2016benchmark} on which the static model is trained. Bottom: directly running the static model, trained on DAVIS without any fine-tuning, on frames from sequences in SegTrackV2 \cite{li2013video}, which contain different object categories. The good segmentation quality in the bottom demonstrates again that our experiential definition of objectness or saliency transfers well between different object categories.}
  \label{fig:vis-transfer}
\end{figure*}

\begin{figure*}[!h]
  \centering
  \includegraphics[width=.99\textwidth]{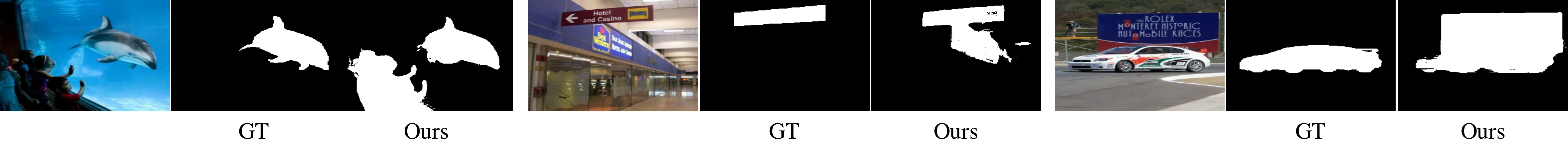}
  \caption{Failure case: multiple salient objects appear in the same image are all captured by our static model. However, the ground-truth annotations only highlight one of them.}
  \label{fig:failure-case}
\end{figure*}

\begin{figure*}[!ht]
  \centering
  \includegraphics[width=0.7\textwidth]{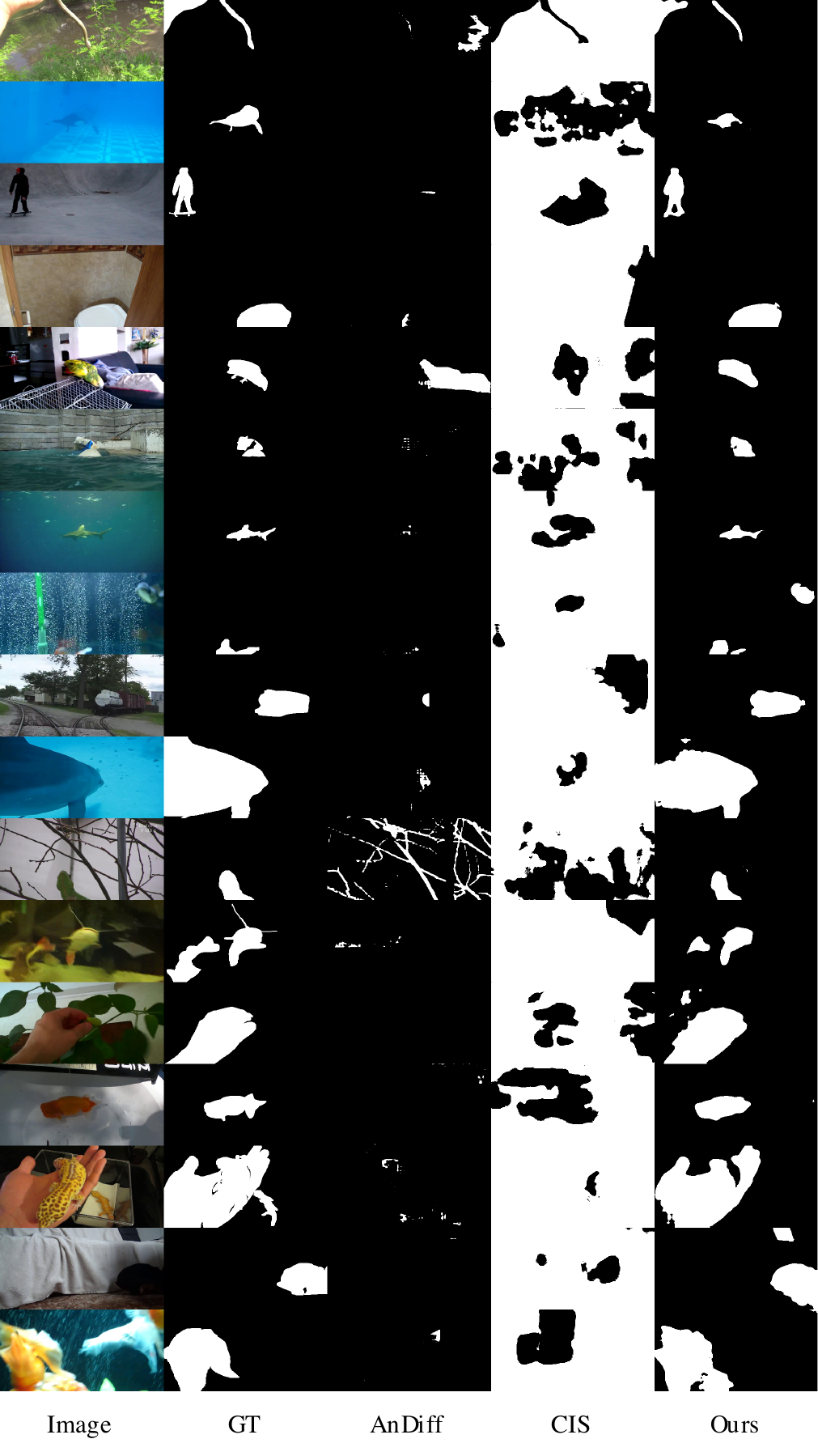}
  \caption{Visual comparison on Youtube-VOS \cite{xu2018youtube}. From left to right: Image, Ground-truth, AnDiff \cite{yang2019anchor}, CIS \cite{yang2019unsupervised}, Ours.}
  \label{fig:yt-vos}
\end{figure*}

\subsection{Additional Qualitative Comparisons}

In Fig.~\ref{fig:vis-transfer}, we show that the objectness prior learned by the static model $\chi$ introduced in Sec.~\ref{subsec:static_object} can effectively generalize to object categories that it has not seen during training. 
We perform the training using only the DAVIS dataset and directly use the static model $\chi$ to do inference on sequences from SegTrackV2.

Youtube-VOS \cite{xu2018youtube} has recently been used to benchmark ``semi-supervised'' video object segmentation methods, 
where one is required to feed the system a ground-truth segmentation mask for the first frame. 
This reduces the problem to one of correspondence and tracking, quite different than what we tackle, which is object 'discovery' without any manual input.
But we still experiment on Youtube-VOS and compare to similar methods to check the generalization ability over diverse video quality, unseen object categories and annotation noise.
We provide additional visual comparisons on the Youtube-VOS \cite{xu2018youtube} benchmark in Fig.~\ref{fig:yt-vos}. In this case, where there are many new categories that have not been seen during training, the model is still capable of generating reasonable segmentations. 
The use of videos to learn good object representations aligns well with the findings in \cite{purushwalkam2020demystifying} which finds that self-supervised contrastive learning methods often perform better when videos are used in place of single images.

\subsection{Failure Case in Salient Object Detection}

In Fig.~\ref{fig:failure-case}, we show some examples of our static model that violate the ground-truth annotations.
Since our model captures the general concept of objectness, it does not have access to or understanding of which object in the image is currently attended by the viewer. Further inclusion of attention information can help on this.

{\small
\bibliographystyle{ieee_fullname}
\bibliography{egbib}
}

\end{document}